\documentclass{article}
\usepackage[T1]{fontenc} 
\usepackage[utf8]{inputenc} 
\usepackage{amsmath,cite,url}
\usepackage{graphicx}
\usepackage[lbd]{ismir}
\usepackage[capitalize]{cleveref}
\usepackage{color}
\usepackage{booktabs} 

\usepackage{amssymb}

\usepackage[firstpage]{draftwatermark}
\definecolor{lightgray}{rgb}{0.9,0.9,0.9}
\definecolor{darkgray}{rgb}{0.4,0.4,0.4}
\SetWatermarkFontSize{12pt}
\SetWatermarkScale{1.1}
\SetWatermarkAngle{90}
\SetWatermarkHorCenter{202mm}
\SetWatermarkVerCenter{170mm}
\SetWatermarkColor{darkgray}
\SetWatermarkText{Late-Breaking / Demo Session Extended Abstract, ISMIR 2025 Conference}

\title{Parametric Neural Amp Modeling with Active Learning}

\multauthor
{Florian Grötschla \hspace{0.5cm} Luca A. Lanzendörfer \hspace{0.5cm} Longxiang Jiao \hspace{0.5cm} \textbf{Roger Wattenhofer}} 
{
ETH Zurich\\
{\tt\small \{fgroetschla, lanzendoerfer, ljiao01, wattenhofer\}@ethz.ch}
}

\def\authorname{F. Grötschla, L. A. Lanzendörfer, L. Jiao, and R. Wattenhofer}

\begin{document}

\newcommand{\model}{\textsc{PANAMA}}

\maketitle
\begin{abstract}
We introduce \model, an active learning framework for the training of end-to-end parametric guitar amp models using a WaveNet-like architecture. With \model, one can create a virtual amp by recording samples that are determined by an active learning strategy to use a minimum amount of datapoints (i.e., amp knob settings). We show that gradient-based optimization algorithms can be used to determine the optimal datapoints to sample, and that the approach helps under a constrained number of samples.   
\end{abstract}
\section{Introduction}\label{sec:introduction}
In recent years, data-driven guitar amp modeling has become increasing popular. Such approaches treat an amp as a blackbox, and simply learn the transform which the amp applies to the raw guitar signal in an end-to-end fashion. The trained models can then be integrated as plugins into Digital Audio Workstations (DAWs), or deployed on modeler pedals. Such a model can be parametric (i.e., conditioned on the amp settings) or non-parametric.
Existing models propose using a feedforward variant of WaveNet or similar Temporal Convolutional Networks (TCNs) for such tasks~\cite{damskägg2019wavenet, Damskgg2019RealTimeMO, Ramirez2020latentWavenet, Steinmetz2021EfficientNN, steinmetz2022efficientneuralnetworksrealtime, Comunit2022ModellingBA}. Other works make use of RNNs frequently paired with LSTM ~\cite{Covert2013AVG, Zhang2018AVG, schmitz2018realtimeemulationparametric, Schmitz2019NonlinearMO, Ramrez2019AGD, wright2019RNN, Wright2020NEURALMO, Wright2021NeuralMO, Sudholt2023PruningDN, Carson2024SampleRI, Yeh2024HyperRN}, and it has further been demonstrated that both architectures can run in real-time on a consumer-grade computer~\cite{wright2020wavenet}. Commercial products like NeuralDSP~\cite{juvela2024endtoendampmodelingdata} also make use of LSTMs, while allowing the model to be conditioned on a full range of amp knob controls. The data collection was automated via a robotic device. Finally, Neural Amp Modeler (NAM)~\cite{NAM} is an open-source non-parametric amp modeler framework, which enables users to capture their own amp setups, and supports different architectures including both a feed-forward WaveNet and RNN.  
To the best of our knowledge, there are currently no open-source parametric amp modelers. The biggest challenge hindering its practicality is arguably the cumbersome data collection procedure, which involves turning the amp knobs to different configurations and recording the amp, where the number of configurations grows exponentially with the number of knobs. We aim to solve this issue using active learning.  
Our main contributions in this paper are: 1) \model ~(\textbf{PA}rametric \textbf{N}eural \textbf{A}mp \textbf{M}odeling with \textbf{A}ctive learning) An open-source parametric amp modeler; 2) An active learning training framework which determines the optimal datapoints to be collected and thus minimizes the total amount needed. To this end, we make use of model ensembling and introduce a method to find the most informative datapoints in a continuous space using gradient-based optimization. 

\section{Methodology}

\begin{figure}
    \centering
    \includegraphics[width=0.5\textwidth]{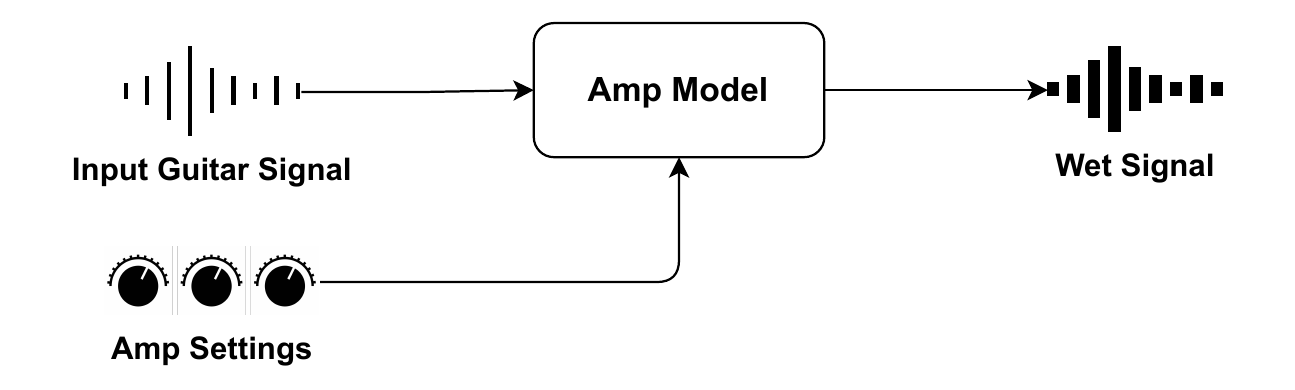}
    \caption{Single Model Setup. The parametric amp model transforms the DI guitar input signal, conditioned on a fixed set of the amp settings.}
    \label{fig:basic}
\end{figure}

\textbf{Model Architecture.}
We adapt the WaveNet~\cite{oord2016wavenetgenerativemodelraw} architecture to be used in a feed-forward manner, similar to prior works. WaveNet was originally developed as an autoregressive generative model, which processes audio using a stack of dilated convolutional layers and predicts a categorical distribution over the next sample to be generated. We use these convolutional layers to transform the input signal, conditioned on amp settings. \cref{fig:basic} provides a high-level visualization.

Let $\mathbf{x}$ be the input guitar signal, $\mathbf{g}$ be a vector containing the value of each amp knob setting as a real number in $[0, 1]$. The original WaveNet architecture provides a way to add both local and global conditioning to the model. The local condition is a time series $\mathbf{y}$, where each timestep is designed to affect the corresponding timestep from the input. Following NAM \cite{NAM}, we set $\mathbf{y} = \mathbf{x}$. The global condition, on the other hand, is a single static vector designed to have effect across all timesteps, which we set to $\mathbf{g}$. Both conditions are incorporated into the convolutional layers as follows: 
\[
z = \tanh(W_{f} * \mathbf{x} + V_{f} * \mathbf{y} + V_{f}'^T \mathbf{g}) \odot \sigma(W_{g} * \mathbf{x} + V_{g} * \mathbf{y} + V_{g}'^T \mathbf{g})
\]
where $W_{*}$ are convolutional kernels, $V_{*}$ are 1x1 convolutional kernels, and $V'_{*}$ are linear mappings. The vector $V'^{T}_{*}\mathbf{g}$ is broadcast over the time dimension.
\begin{figure}
    \centering
    \includegraphics[width=0.5\textwidth]{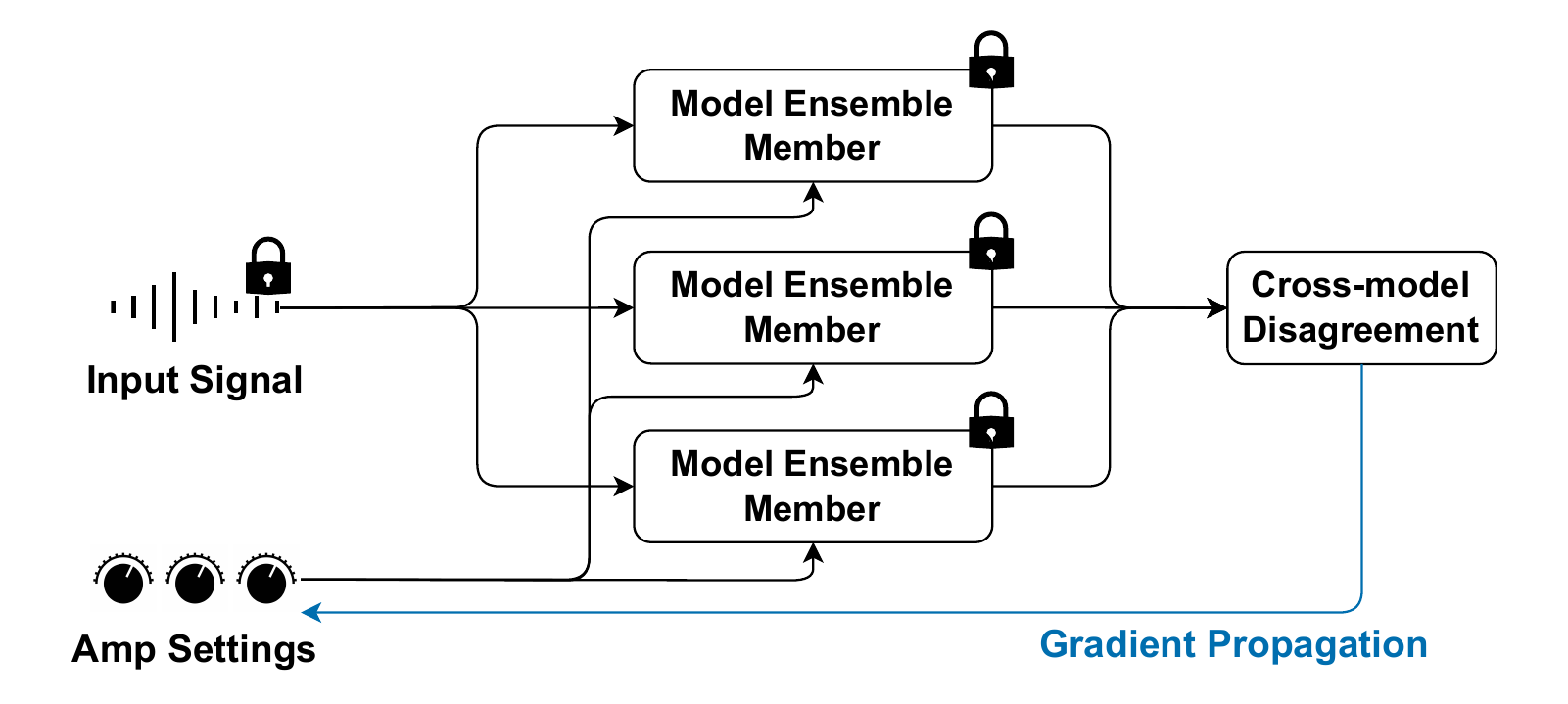}
    \caption{Active Learning Setup. The input signal $\mathbf{x}$ and amp settings $\mathbf{g}$ are fed into independently trained instances of the model, and variation in the outputs is calculated as the cross-model disagreement $D$. Gradients are propagated back to $\mathbf{g}$ in order to maximize $D$.}
    \label{fig:actlearn}
\end{figure}
\begin{figure}[b]
    \centering
    \includegraphics[width=0.45\textwidth]{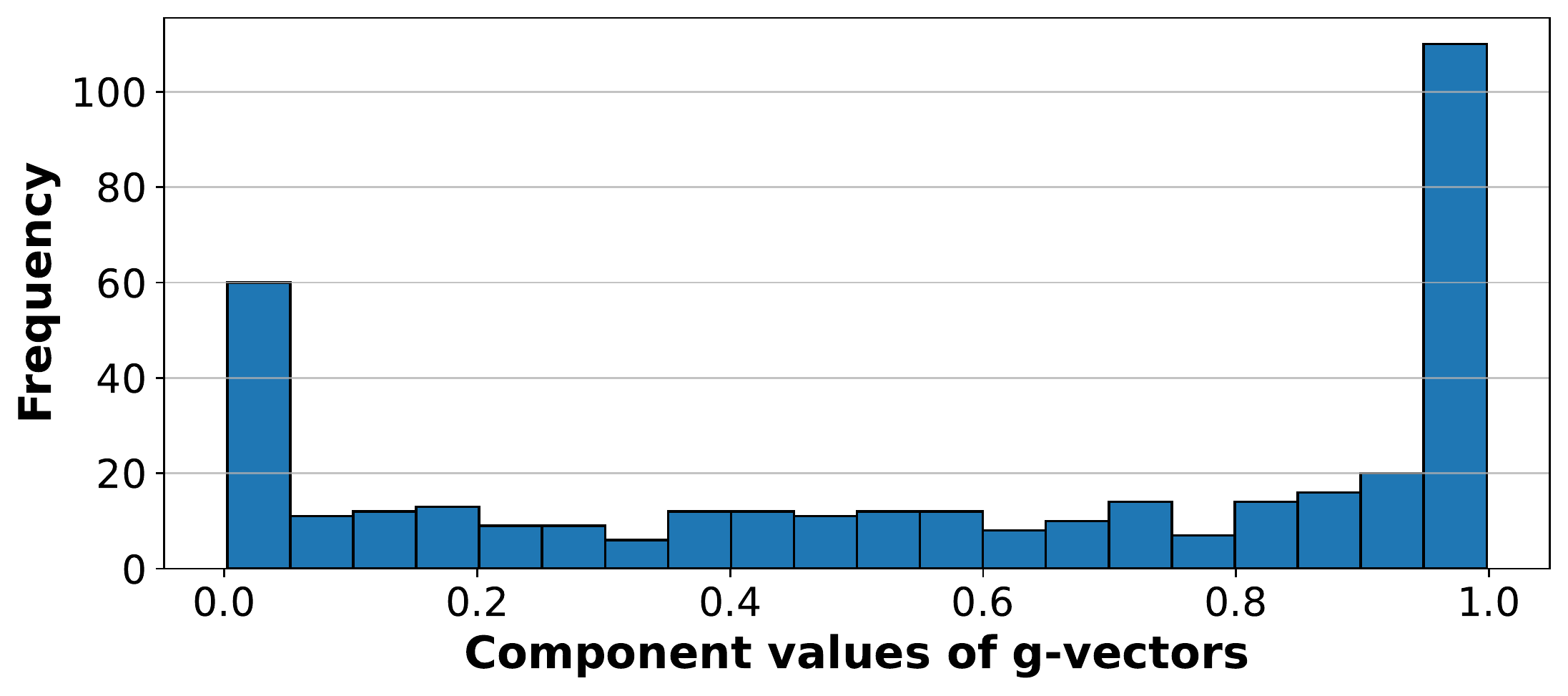}
    \caption{Distribution of component values in $\mathbf{g}$-vectors gathered through active learning.}
    \label{fig:g_dist}
\end{figure}

\noindent\textbf{Active Learning Approach.}
We now take the following perspective: An input signal $\mathbf{x}$ and an amp settings vector $\mathbf{g}$ together constitute an unlabeled datapoint, whereas their corresponding wet signal is the ground truth label. A dataset consists of such datapoint-label pairs. The goal of our active learning method is to find the optimal next datapoint to ``label'' (which involves running the fixed input signal through the amp with the given settings, and recording the output audio), given the current set of already labeled points $\mathbf{G}$. 
Typically in active learning, we would choose the datapoint that maximizes the uncertainty of the model, as this means the new point would provide the most information. However, here we cannot compute the uncertainty directly and opt to use model disagreement over the datapoint across an ensemble, denoted as $D$, as a proxy.
Suppose that we have a model ensemble $\mathcal{F}_\mathbf{G}$ of size $M$, where $f^{(i)}_\mathbf{G} \in \mathcal{F}_\mathbf{G}$ denotes the $i$-th model. Each $f^{(i)}_\mathbf{G}$ is trained independently with the current dataset $\mathbf{G}$, using random initialization, dataset shuffling etc. so that they converge to slightly different results. Then we have:
\[
D_\mathbf{G}(\mathbf{x, g}) = \frac{1}{M}\operatorname{tr}(\operatorname{Var}_i[f^{(i)}_\mathbf{G}(\mathbf{x, g})]),
\]
where $\operatorname{Var}$ denotes the covariance matrix. In other words, we take the cross-model variances of the output signals in element and then average them into one scalar. 
We then keep $\mathbf{x}$ as a pre-chosen, fixed signal, and define the optimal $\mathbf{g}$ as:
\[
\mathbf{g^{*}} := \operatorname{argmax}_{\mathbf{g} \in \mathbb{R}^k} D_\mathbf{G}(\mathbf{x, g}),
\]
where $k$ is the number of settings, or knobs on the amp to be modeled.
Importantly, since the calculation of model disagreement is differentiable, we can propagate the gradients from $D_\mathbf{G}$ all the way back to $\mathbf{g}$. Thus, we can simply use any gradient-based optimization algorithm to find $\mathbf{g^{*}}$.
An overview of the approach can be found in \cref{fig:actlearn}.

\section{Experimental Evaluation}

For our experiments, we choose $\operatorname{dim}(\mathbf{g}) = 6$. This includes the following parameters: Gain, Bass, Mid, Treble, Master, Presence. The Master knob is included, as it differs from a standard volume knob. The fixed input signal is taken from NAM~\cite{NAM} and roughly 3 minutes long. We obtain the ground truth signals from an amp sim. All models are trained for 50 epochs with MSE loss. For validation, we use 30 minutes of guitar audio across different genres, gathered from the IDMT-SMT-GUITAR dataset \cite{kehling_2023_idmt_smt_guitar}, and around 1,000 randomly sampled amp settings. The code is available online~\footnote{\scriptsize{\url{https://github.com/ETH-DISCO/neural_amp_modelling}}}.
Training a single model with 300 uniformly randomly chosen datapoints, we are able to achieve an MSE of 7e-05 on the validation set.
For active learning, We use a model ensemble of size 4 and start with 10 randomly chosen points. In each round, each model in the ensemble is first trained on the currently gathered datapoints. Then, an Adam optimizer is used to maximize the cross-model disagreement 10 independent times. From the resulting 10 local optima, we extract the unique ones using a clustering algorithm. This usually yields 4-5 different $\mathbf{g}$-vectors. We stop after reaching 64 datapoints in total. 
The chosen vectors tend to have values near 0 or 1. This aligns with the intuition that knob settings close to the min/max positions are harder for the model to obtain through interpolation of seen data. We visualize the component values of gathered $\mathbf{g}$-vectors (by flattening all vectors into one list of scalars) as a histogram in \cref{fig:g_dist}. 
\begin{table}[t]
    \centering
    \begin{tabular}{lc}
    \toprule
    \textbf{Sampling Method} & \textbf{Val. Set MSE} \\
    \midrule
    Uniform Random & 8.6e-04\\
    Beta Distr. ($\alpha=\beta=0.5$) & 9e-04 \\
    \model & 3.4e-04 \\
    \bottomrule
    \end{tabular}
    \caption{Comparison of different methods to sample 64 datapoints. Active learning performs best.}
    \label{tab:comparison}
\end{table}
We also observe that this histogram resembles a beta distribution, and fitting it to one yields $\alpha = 0.5396, \beta = 0.4122$. Based on this, we conduct an additional experiment, sampling data points purely from a beta distribution with $\alpha = \beta = 0.5$ as a heuristic for comparison. 
\cref{tab:comparison} compares all sampling approaches using 64 datapoints. The beta heuristic performs slightly worse than random sampling, while active learning yields a significantly lower validation error. This suggests that, although active learning favors extreme values, its behavior is more complex than the heuristic.

\bibliography{ISMIRtemplate}

%
%
%
%
%

\end{document}